\documentclass[letterpaper,journal]{IEEEtran}

\usepackage[utf8]{inputenc}
\usepackage[T1]{fontenc}
\usepackage{amsmath,amssymb,amsfonts}
\usepackage{mathrsfs,bbm}
\usepackage{array}
\usepackage{booktabs}
\usepackage{cite}
\usepackage{graphicx}
\usepackage{multirow}
\usepackage{placeins}
\usepackage{stfloats}
\usepackage{textcomp}
\usepackage{url}
\usepackage{verbatim}
\usepackage[table]{xcolor}
\usepackage{nicefrac}
\usepackage{microtype}
\usepackage{multicol}
\usepackage{wrapfig}
\usepackage{algorithm}
\usepackage{algpseudocode}
\usepackage{hyperref}
\usepackage[capitalize]{cleveref}
\hypersetup{hidelinks}

\crefname{section}{Sec.}{Secs.}
\Crefname{section}{Section}{Sections}
\crefname{figure}{Fig.}{Figs.}
\Crefname{figure}{Fig.}{Figs.}
\crefname{table}{Table}{Tables}
\Crefname{table}{Table}{Tables}


\title{Physical SLAM: Physics-aware Simultaneous Localisation and Mapping}

\author{\mbox{Guangming Wang}, \mbox{Zhishuo Liu}, \mbox{Haibing Wu}, \mbox{Wanli Ma},
\mbox{Yixiong Jing}, \mbox{Olaf Wysocki}, and \mbox{Brian Sheil}%
\thanks{All authors are with the University of Cambridge, Cambridge, U.K. (e-mail: \{gw462,zl678,hw657,wm369,yj401,okw24,bbs24\}@cam.ac.uk).}%
\thanks{G. Wang and Z. Liu contributed equally. Corresponding author: Haibing Wu (e-mail: hw657@cam.ac.uk).}}

\begin{document}
\bstctlcite{BSTcontrol}

\maketitle


\begin{abstract}
To reconstruct a predictive 4D representation of the physical world from monocular video captured by a moving camera, we jointly reconstruct dynamic Gaussian radiance fields and infer their underlying physical dynamics. This paper proposes \emph{Physical SLAM}, a physics-aware method for constructing 4D Gaussian fields of deformable objects and predicting their future state when subjected to external actions, given a sequence of RGB-D images. Our method separates the scene into a static background Gaussian field for camera localization and a dynamic object Gaussian field driven by a spring--mass physical model of elastic deformation, inertia, and damping. To integrate visual observations with physical dynamics, we introduce virtual observation forces that gradually pull simulated surface nodes toward observed depth points, avoiding abrupt replacement of simulated node positions. Meanwhile, the physical parameters of the spring--mass model are updated from simulation errors aggregated over the full reconstruction observation window, enabling the reconstructed dynamics to align with observations. To couple appearance, geometry, and physics, we further design a physics-aware Gaussian transformation and surface alignment strategy, where dynamic Gaussians are transformed along the deforming physical surface of the spring--mass model and constrained to remain aligned with the surface during rendering-based optimization. The resulting representation supports observed-frame rendering and action-conditioned simulation, extending SLAM maps to describe how deformable objects respond to manipulation. On the PhysTwin benchmark, our method improves reconstruction and prediction PSNR over PhysTwin by 10.4\% and 11.0\%, respectively. On the TUM RGB-D benchmark, it reduces average camera tracking error by 20.6\%, improves SSIM by 3.6\%, and reduces LPIPS by 8.3\% compared with 4DGS-SLAM. The source code will be released upon acceptance.

        %
\end{abstract}
\section{Introduction}
\label{sec:intro}

\IEEEPARstart{T}{racking}, mapping, rendering, and predicting dynamic 3D scenes from streaming RGB-D observations is a fundamental problem in computer vision and robotics. SLAM estimates camera motion and scene structure from sequential observations~\cite{teed2021droid}. Geometric maps support localisation, while rendering and physical interaction additionally require appearance and dynamics~\cite{matsuki2024gaussian,jiang2025phystwin}.
3D Gaussian Splatting (3DGS) has recently emerged as an efficient explicit representation for high-fidelity scene reconstruction and rendering~\cite{kerbl20233d}.
 Gaussian-based SLAM systems have leveraged this representation to incrementally estimate camera poses and optimize Gaussian primitives from RGB or RGB-D streams, achieving accurate localization and high-quality dense mapping in unknown environments~\cite{yan2024gsslam,matsuki2024gaussian,keetha2024splatam,huang2024photoslam,yugay2023gaussianslam}.
However, previous methods have been primarily designed for static scenes.
When moving or deformable objects appear, a static Gaussian map can introduce inconsistent geometry, blurred regions, and degraded camera tracking thereby inhibiting its application to dynamic problems~\cite{bescos2018dynaslam,li20254dgsslam}.

To handle non-static environments, many dynamic SLAM and Gaussian SLAM methods detect moving regions and remove, mask, or down-weight them during tracking and mapping~\cite{bescos2018dynaslam,xu2024dgslam,kong2024dgslam,wen2024gassidy,zheng2025wildgs,yang2025bdgs}. 
Although this strategy improves camera localization and produces cleaner static reconstructions, dynamic objects are treated as distractors rather than reconstructed as part of the 4D world.
Recent 4D Gaussian Splatting SLAM methods bridge this gap by explicitly maintaining dynamic Gaussian primitives and modeling their temporal transformations~\cite{li20254dgsslam,matsuki20254dtam}.
Instead of discarding moving regions, they reconstruct time-varying Gaussian radiance fields and enable streaming reconstruction of dynamic appearance and geometry. Despite this progress, current 4D Gaussian fields are still primarily visual and kinematic.
Their temporal evolution is usually parameterized by deformation fields~\cite{park2021nerfies,pumarola2021dnerf,yang2024deformable3dgs}, sparse control points~\cite{huang2024scgs,li20254dgsslam}, or learned transformations ~\cite{wu20244dgs,bae2024pergaussian,sun20243dgstream}, and is often further supervised with optical-flow constraints~\cite{wang2023flow,li20254dgsslam,teed2020raft}.
Such representations can describe where dynamic Gaussians move in observed frames, but they do not explain why the object deforms or how the object will respond to future actions.
This missing physical layer is critical for robotic manipulation \cite{jiang2025phystwin}.

\begin{figure*}[t]
    \centering
\includegraphics[width=1.0\textwidth]{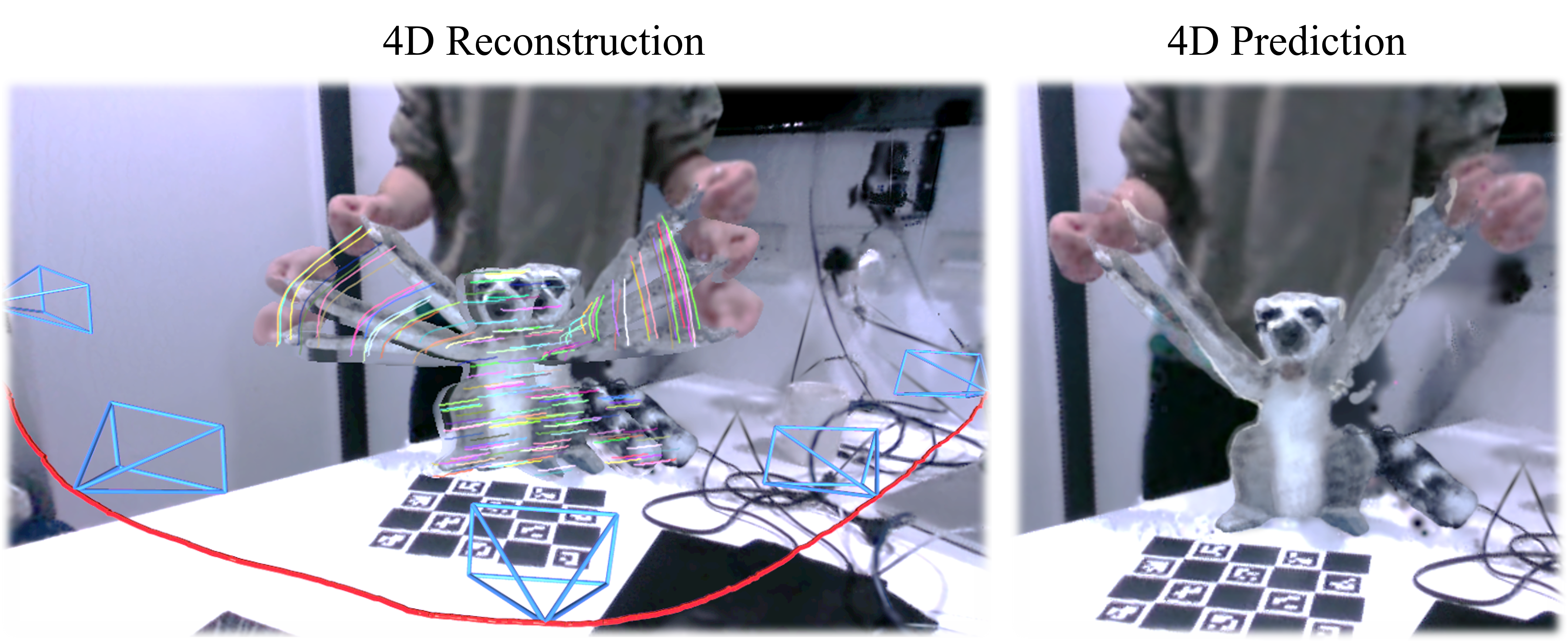}
    \caption{
    \textbf{Physics-aware SLAM with 4D Gaussian field for reconstruction and prediction.}
    Existing 4D Gaussian SLAM methods mainly reconstruct time-varying geometry and appearance. Our method further reconstructs the physical dynamics of deformable objects, enabling online reconstruction and physically grounded action-conditioned prediction from RGB-D observations. For prediction, our method simulates the dynamics of the spring--mass model under the supplied actions and renders the resulting object states from a query camera pose.}
    \label{fig:p1}
\end{figure*}

Recent physics-aware reconstruction methods show that incorporating physical models into object reconstruction can improve deformable object simulation and action-conditioned future prediction~\cite{zhang2024physdreamer,zhong2024springgaus,jiang2025phystwin}.
However, these methods are mostly designed for offline reconstruction of deformable object geometry, appearance, and dynamics from previously collected observations.
They do not reconstruct a physics-aware 4D Gaussian field in a streaming setting, where camera poses, appearance, geometry, and physical dynamics are updated from incoming RGB-D frames. The remaining challenge is to combine camera localisation with progressive estimation of deformable geometry, appearance, and physical dynamics in a single RGB-D pipeline.

To bridge this gap, we introduce \emph{Physical SLAM}, a method for reconstructing physics-aware 4D Gaussian fields from a single RGB-D stream, as shown in Fig.~\ref{fig:p1}.
Given an RGB-D sequence, we separate the scene into a static background Gaussian field and a dynamic object Gaussian field similar to \cite{li20254dgsslam}.
The static field supports camera localization and background mapping, while the deformable object is represented as surface-bound dynamic Gaussians coupled with a spring--mass physical model.
Rather than modeling dynamic Gaussians with purely kinematic transformations, the object's physical state is predicted by the physical model combined with incoming RGB-D observations through our proposed per-node observation forces, and then used to drive the motion of dynamic Gaussians along the deforming surface.
During mapping, the system jointly refines the background field, the physical parameters, and the surface-bound object Gaussians from RGB-D observations.
The reconstructed 4D Gaussian field supports rendering at observed frames and predicts object deformation and the corresponding Gaussian motion under external actions.

The novel contributions of our method are summarized as follows:
\begin{itemize}

    \item We propose a physics-aware 4D Gaussian reconstruction method that couples dynamic Gaussian primitives with a spring--mass physical model, enabling joint reconstruction of appearance, geometry, and physical dynamics from streaming RGB-D observations.

    \item We introduce an observation-force-based physical tracking formulation that injects visual observations into the dynamics model through virtual forces, enabling stable deformable object tracking without direct state replacement.

    \item We propose a physics-aware Gaussian transformation with a streaming optimization scheme for dynamic Gaussian primitives, allowing the reconstructed 4D Gaussian field to support both high-fidelity rendering and physically grounded action-conditioned prediction.
\end{itemize}

\section{Related Work}
\label{sec:related}

\noindent\textbf{3D Gaussian Splatting and Non-static GS SLAM.}
3DGS ~\cite{kerbl20233d} provides an explicit and efficient radiance-field representation for rendering and reconstruction, and has been extended to geometry-aware reconstruction and dense SLAM from monocular, stereo, and RGB-D streams~\cite{kerbl20233d,yan2024gsslam,matsuki2024gaussian,keetha2024splatam,huang2024photoslam,yugay2023gaussianslam}.
These methods achieve strong tracking and mapping performance in static or mostly rigid scenes, but dynamic objects can cause pose errors and ghosting artifacts.
To improve robustness in non-static environments, dynamic SLAM and non-static GS-SLAM methods ~\cite{bescos2018dynaslam,jiang2024rodynslam,xu2024dgslam,kong2024dgslam,wen2024gassidy,zheng2025wildgs,yang2025bdgs} detect dynamic or disturbance regions through various methods, such as motion masks~\cite{xu2024dgslam}, rendering residuals ~\cite{wen2024gassidy}, uncertainty maps~\cite{zheng2025wildgs}, or probabilistic Gaussian labels ~\cite{yang2025bdgs}, to suppress dynamic distractors during pose estimation and reconstruct cleaner static Gaussian fields. 

\noindent\textbf{Dynamic Gaussian Splatting.}
Dynamic neural rendering and dynamic Gaussian Splatting extend static radiance-field reconstruction to time-varying scenes through deformation fields~\cite{park2021nerfies,pumarola2021dnerf,matsuki20254dtam}, factorized 4D feature planes~\cite{cao2023hexplane}, 4D neural voxels and deformation MLPs~\cite{wu20244dgs}, canonical-space Gaussian deformation~\cite{yang2024deformable3dgs}, persistent Gaussian trajectories~\cite{luiten2024dynamic3dgs}, sparse control points~\cite{huang2024scgs,li20254dgsslam}, neural transformation caches~\cite{sun20243dgstream}, and per-Gaussian embeddings~\cite{bae2024pergaussian}.
These methods significantly improve dynamic novel-view synthesis by learning how Gaussian primitives or radiance fields move across time.
More closely related to our setting, 4D Gaussian Splatting SLAM separates static and dynamic Gaussians and learns dynamic transformations with sparse control points and an MLP~\cite{li20254dgsslam}. 
These kinematic transformations do not model the forces and material dynamics needed for action-conditioned prediction. Our method addresses this limitation in the dynamic branch. We bind dynamic Gaussians to a spring--mass model and update the physical state using our proposed virtual observation forces. 
The reconstructed 4D Gaussian field captures time-varying geometry, appearance, and physical dynamics for action-conditioned future prediction.

\noindent\textbf{Physics-aware deformable modeling and action-conditioned prediction.}
Beyond purely visual 4D reconstruction, physics-aware methods seek to build simulatable object representations by coupling visual observations with physical dynamics.
Existing approaches instantiate this idea with different scene representations, including point-cloud/mesh-based deformable modeling and differentiable simulation~\cite{wang2015deformation,jatavallabhula2021gradsim,du2021diffpd}, SDF-based physical reconstruction~\cite{qiao2022neuphysics}, NeRF or neural-field-based simulatable objects~\cite{chen2022virtualelasticobjects,li2023pacnerf,feng2024pienerf}, and Gaussian-based physical representations~\cite{xie2024physgaussian,zhang2024physdreamer,zhong2024springgaus,jiang2025phystwin}. Some of them prescribe material parameters or user-defined physical settings~\cite{feng2024pienerf,xie2024physgaussian}, while others perform inverse identification from videos or observations~\cite{chen2022virtualelasticobjects,li2023pacnerf,qiao2022neuphysics,zhang2024physdreamer,zhong2024springgaus,jiang2025phystwin}.
PhysTwin is the closest work to ours, as it combines spring-mass dynamics, Gaussian splats, and inverse physics optimization for deformable digital-twin reconstruction~\cite{jiang2025phystwin}.
PhysTwin reconstructs objects from three fixed-view RGB-D recordings in an offline pipeline. Our method estimates camera poses and builds a physics-aware 4D Gaussian field online from a single moving RGB-D stream, updating the map as frames arrive.
Moreover, learning-based dynamics models~\cite{sanchezgonzalez2020learning,pfaff2021learning,zhang2025gsdynamics} require extensive training samples and are often limited to specific environments, whereas our method does not require pre-training and establishes physical understanding during the SLAM process.

\section{Preliminary}
\label{sec:Preliminary}

\noindent\textbf{3D Gaussian Splatting.}
We use 3DGS as the explicit scene representation for rendering and optimisation~\cite{kerbl20233d}.
A scene is represented by a set of anisotropic Gaussian primitives $\mathcal{G}=\{g_i\}_{i=1}^{N}$.
Each Gaussian $g_i$ is parameterized by its center $\boldsymbol{\mu}_i\in\mathbb{R}^{3}$, covariance $\boldsymbol{\Sigma}_i\in\mathbb{R}^{3\times 3}$, opacity $o_i\in[0,1]$, and a view-independent RGB color $\mathbf{c}_i$, where spherical-harmonic coefficients are omitted for efficient optimization following recent Gaussian SLAM systems~\cite{matsuki2024gaussian,yugay2023gaussianslam}.
The covariance is factorized as $\boldsymbol{\Sigma}_i=\mathbf{R}_i\operatorname{diag}(\mathbf{s}_i^2)\mathbf{R}_i^{\top}$, where $\mathbf{R}_i$ and $\mathbf{s}_i$ denote the rotation and scale of the Gaussian. 

Given a camera pose $\mathbf{T}_{wc}$, we project each Gaussian onto the image plane.
Let $\mathbf{T}_{cw}=\mathbf{T}_{wc}^{-1}$ and $\Pi(\cdot)$ be the camera projection function.
The projected mean and covariance are $\boldsymbol{\mu}'_i=\Pi(\mathbf{T}_{cw}\boldsymbol{\mu}_i)$ and $\boldsymbol{\Sigma}'_i=\mathbf{J}_i\mathbf{R}_{cw}\boldsymbol{\Sigma}_i\mathbf{R}_{cw}^{\top}\mathbf{J}_i^{\top}$, where $\mathbf{J}_i$ is the Jacobian of the local projective transformation and $\mathbf{R}_{cw}$ is the rotation component of $\mathbf{T}_{cw}$.
For a pixel $p$, the density of Gaussian $g_i$ is $\alpha_i(p)=o_i\exp\!\left(-\frac{1}{2}(p-\boldsymbol{\mu}'_i)^{\top}{\boldsymbol{\Sigma}'_i}^{-1}(p-\boldsymbol{\mu}'_i)\right)$.
With transmittance $T_i(p)=\prod_{j<i}(1-\alpha_j(p))$ and blending weight $w_i(p)=T_i(p)\alpha_i(p)$, the rendered color and depth are computed as $\hat{I}(p)=\sum_{i\in\mathcal{N}(p)}w_i(p)\mathbf{c}_i$ and $\hat{Z}(p)=\sum_{i\in\mathcal{N}(p)}w_i(p)z_i$, where $z_i$ is the depth of the Gaussian center and $\mathcal{N}(p)$ denotes the ordered set of Gaussians overlapping pixel $p$.
This differentiable rendering process allows Gaussian parameters and camera poses to be optimized using photometric and geometric losses.

\noindent\textbf{Spring--Mass based Reconstruction and Simulation of Deformable Objects from Videos.}
A deformable object can be represented as a spring--mass graph $\mathcal{H}=(\mathcal{V},\mathcal{E})$, where $\mathcal{V}$ denotes mass nodes and $\mathcal{E}$ denotes spring connections.
The state at time $t$ is $\mathbf{X}_t=\{(\mathbf{x}_i^t,\mathbf{v}_i^t)\}_{i=1}^{N_s}$, and the dynamics are parameterized by $\boldsymbol{\theta}=\{m_i,k_{ij},\gamma_{ij},\ell_{ij}\}$, including node masses, spring stiffness, damping coefficients, and rest lengths.
For each node $i$, let 
$\mathcal{N}_{\mathcal{H}}(i)=\{j\mid (i,j)\in\mathcal{E}\ \text{or}\ (j,i)\in\mathcal{E}\}$ 
denote its neighboring nodes in the spring graph.
The total force is commonly written as
\begin{equation}
    \mathbf{F}_i
    =
    \sum_{j\in\mathcal{N}_{\mathcal{H}}(i)}
    \left(
    \mathbf{F}_{ij}^{\rm spring}
    +
    \mathbf{F}_{ij}^{\rm damp}
    \right)
    +
    \mathbf{F}_{i}^{\rm ext},
    \label{force}
\end{equation}
where $\mathbf{F}_{ij}^{\rm spring}=k_{ij}(\|\mathbf{x}_j-\mathbf{x}_i\|-\ell_{ij})\mathbf{e}_{ij}$, $\mathbf{F}_{ij}^{\rm damp}=\gamma_{ij}((\mathbf{v}_j-\mathbf{v}_i)^{\top}\mathbf{e}_{ij})\mathbf{e}_{ij}$, and $\mathbf{e}_{ij}=(\mathbf{x}_j-\mathbf{x}_i)/\|\mathbf{x}_j-\mathbf{x}_i\|$.
Given an action signal $\mathbf{a}_t$, the simulator evolves the state as $\mathbf{X}_{t+1}=f_{\boldsymbol{\theta},\mathcal{H}}(\mathbf{X}_t,\mathbf{a}_t)$.

Spring--mass-based reconstruction from videos can be viewed as inverse modeling.
Given an RGB-D observation $\mathcal{O}_t=(I_t,Z_t)$ and camera pose $\mathbf{T}_{wc}^t$, the predicted observation is 
$\hat{\mathcal{O}}_t=\mathcal{R}_{\boldsymbol{\psi}}(\mathbf{X}_t,\mathbf{T}_{wc}^t)$.
The physical and appearance parameters are estimated by solving
$\min_{\boldsymbol{\theta},\boldsymbol{\psi}}\sum_t\mathcal{C}(\hat{\mathcal{O}}_t,\mathcal{O}_t)$.
The cost typically combines geometry, motion, and rendering terms, i.e., $\mathcal{C}=\mathcal{C}_{\rm geo}+\mathcal{C}_{\rm motion}+\mathcal{C}_{\rm render}$.
PhysTwin~\cite{jiang2025phystwin} uses this formulation to reconstruct deformable 3D object models with geometry, appearance, and physical parameters from interaction videos.
Our method adopts the same spring--mass principle as the physical backbone, but integrates it into a streaming 4D Gaussian reconstruction system with observation-force correction, progressive parameter updating, and physics-aware Gaussian transformation with constrained progressive optimization.

\begin{figure*}[t]
    \centering
    \includegraphics[width=\textwidth]{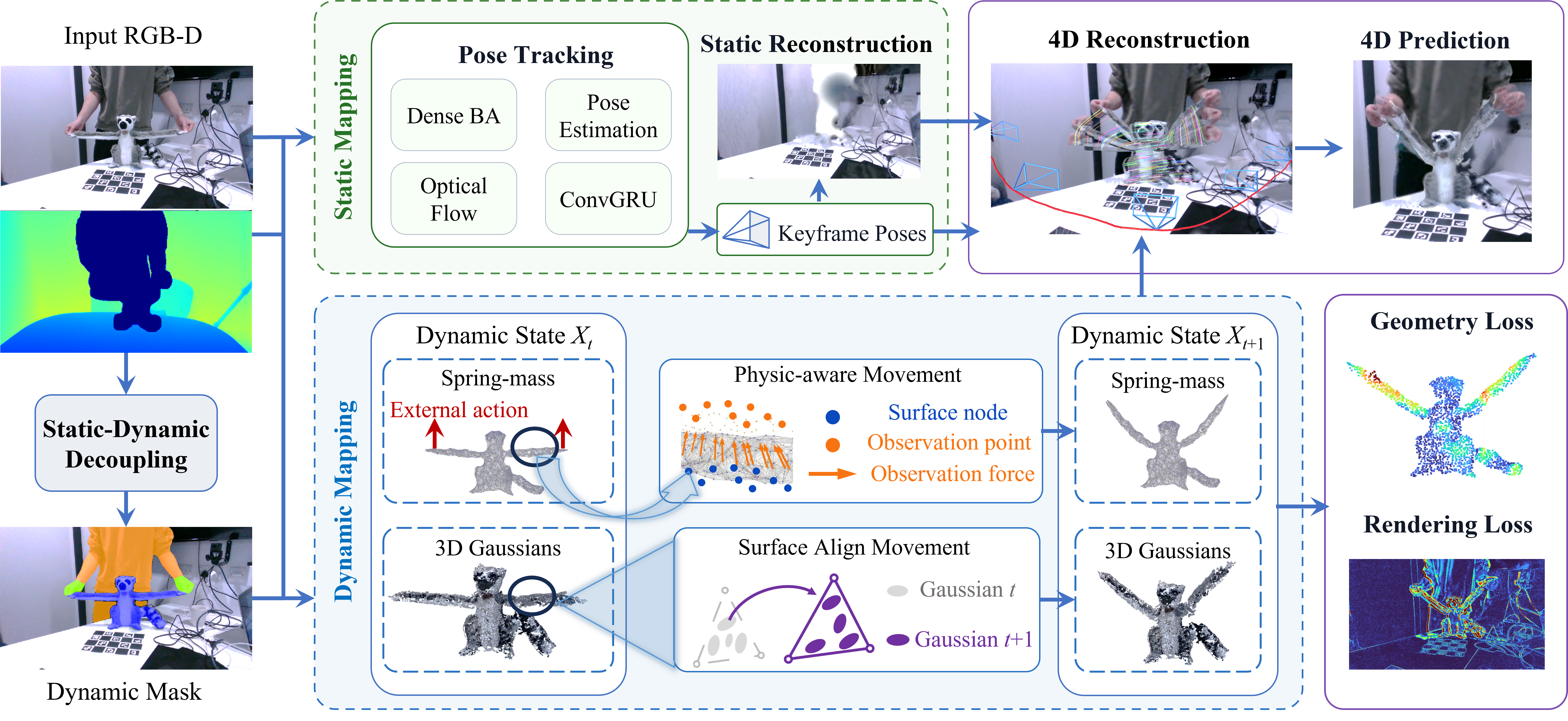}
    \vspace{-6mm}
    \caption{\textbf{Pipeline of Physical SLAM.} We decouple the static and dynamic mapping using semantics. For the static part, we build a globally consistent map and use the map to track the camera pose. For the dynamics part, we use a spring--mass system with external actions and observation forces to track the trajectory of surface nodes and use the geometry loss to progressively optimize the physical parameters; in parallel, we use a surface alignment method to align the 3D Gaussians to the mesh of the object to realise the tightly coupled optimization of physical dynamics, geometry, and appearance.}
    \label{fig:sfc}
\end{figure*}

\section{Method}
\label{sec:method}


Given a single-camera RGB-D stream $\mathcal{O}_{1:T}=\{(I_t,Z_t)\}_{t=1}^{T}$, our goal is to reconstruct a physics-aware 4D Gaussian field that supports both observed-frame rendering and action-conditioned future prediction.
As shown in Fig.~\ref{fig:sfc}, we decompose the scene into a static background field $\mathcal{G}^{\rm bg}$ and a physics-aware dynamic object field $\mathcal{G}^{\rm obj}_t$, following recent dynamic Gaussian SLAM systems~\cite{li20254dgsslam}.
For each frame, a pre-trained model YoLov9 ~\cite{wang2024yolov9} provides the object mask $M_t^{\rm obj}$ and background mask $M_t^{\rm bg}=1-M_t^{\rm obj}$, same to \cite{li20254dgsslam}.
Background RGB-D pixels initialize and expand $\mathcal{G}^{\rm bg}$, which is used for camera tracking and static mapping.
For the evaluated sequences, we use separate initialization procedures for active moving objects (like humans) and passive moving objects (like the manipulated deformable objects). Currently we have only tested the "human" type of active moving objects. For humans, the PHALP~\cite{rajasegaran2022tracking} representation consists of meshes generated by the SMPL body model~\cite{loper2015smpl}. Mesh association across frames is obtained primarily from the tracking pipeline of 4DHumans~\cite{goel2023humans}. For the manipulated objects, following PhysTwin~\cite{jiang2025phystwin}, we initialize a canonical mesh $\mathcal{M}_0$ from the first-frame foreground using TRELLIS~\cite{xiang2024trellis}, align it to the observed RGB-D geometry, and construct a spring graph $\mathcal{H}=(\mathcal{V},\mathcal{E})$ by connecting nearby sampled surface nodes. At time $t$, the object surface is represented as $\mathcal{M}_t=(\mathbf{P}_t,\mathcal{F})$, where $\mathbf{P}_t=\{\mathbf{p}_i^t\}$ are vertices driven by the corrected physical state $\hat{\mathbf{X}}_t$ and $\mathcal{F}$ is the fixed face connectivity.
The state $\hat{\mathbf{X}}_t$ is corrected by observation forces in Sec.~\ref{sec:tracking}, the physical parameters $\boldsymbol{\theta}_t$ are updated in Sec.~\ref{sec:physics_4d_mapping}. Object pixels initialize the canonical object Gaussian field $\mathcal{G}^{\rm obj}_{\rm c}$, which is bound with the mesh as follows.

\noindent\textbf{Dynamic Gaussian initialization with physical surface binding.}
Unlike 4DGS-SLAM, which learns kinematic transformations for dynamic Gaussians, we bind each object Gaussian $g_n\in\mathcal{G}^{\rm obj}_{\rm c}$ to the physical surface.
Specifically, $g_n$ is attached to the closest canonical mesh face $f_{q(n)}=(\mathbf{p}_{q(n),1}^{0},\mathbf{p}_{q(n),2}^{0},\mathbf{p}_{q(n),3}^{0})$ by storing its barycentric coordinate and normal offset:
\begin{equation}
    \boldsymbol{\mu}_n^0
    =
    \sum_{r=1}^{3}
    \beta_{nr}\mathbf{p}_{q(n),r}^{0}
    +
    h_n\mathbf{n}_{q(n)}^{0},
    \qquad
    \sum_{r=1}^{3}\beta_{nr}=1.
    \label{eq:gaussian_binding}
\end{equation}
Here $\boldsymbol{\beta}_n$ is the barycentric coordinate, $h_n$ is the signed offset along the face normal, and $\mathbf{n}_{q(n)}^{0}$ is the normal of the canonical face.
The binding set $\mathcal{B}=\{(q(n),\boldsymbol{\beta}_n,h_n)\}_{g_n\in\mathcal{G}^{\rm obj}_{\rm c}}$ defines the physics-aware transformation used for dynamic Gaussian transformation in Sec.~\ref{sec:physics_4d_mapping}.

\subsection{Tracking with Observation-Force-Corrected Dynamics}
\label{sec:tracking}

\noindent\textbf{Static camera tracking.} 
Our camera tracking follows the DROID-SLAM-style dense bundle adjustment (DBA) backend adopted by WildGS-SLAM~\cite{teed2021droid,zheng2025wildgs}.
Unlike WildGS-SLAM, which learns a per-pixel uncertainty map to down-weight dynamic distractors, we directly use the semantic dynamic mask $M_t^{\rm obj}$. This provides an explicit static/dynamic partition shared by camera tracking and object reconstruction.
The tracking backend optimizes camera poses from RGB-D and optical-flow constraints only on background pixels, and passes the estimated keyframe poses to the mapping module.

\noindent\textbf{Tracking dynamics with observation-force correction.}
Given the corrected state $\hat{\mathbf{X}}_{t-1}$, we advance the node positions and velocities with the spring--mass simulator defined in Sec.~\ref{sec:Preliminary}.
When explicit manipulation is available, we track points in segmented contact regions with CoTracker3~\cite{karaev2025cotracker3} and lift them to 3D using depth and camera poses. Their trajectories define the action $\mathbf{a}_{t-1}$ and drive nearby object nodes through control springs, following PhysTwin~\cite{jiang2025phystwin}.
For dynamic objects with unknown actions, such as human foregrounds in generic RGB-D SLAM sequences, we set $\mathbf{a}_{t-1}=\mathbf{0}$ and rely on observation-force correction for tracking.
A pure action-driven rollout accumulates errors due to noisy actions and imperfect physical parameters, while directly replacing simulated nodes with observed geometry points may introduce discontinuities and unstable oscillations. We therefore convert geometry observations into external observation forces.

Let $\mathcal{D}_t=\{\mathbf{d}_j^t\}$ be the object point cloud back-projected from $M_t^{\rm obj}$.
At simulation substep $s$, each node $\mathbf{x}_i^{(s)}$ is projected to the current depth image, and its visibility ${\rm vis}_i^{(s)}$ is determined by a depth-based occlusion test.
A node is visible if it is in front of the camera, projects inside the image with valid depth, and is not behind the observed surface beyond a depth tolerance $\epsilon_z>0$. Only visible nodes receive observation forces.
For each visible node $\mathbf{x}_i^{(s)}$, we collect nearby observed points in the set $\mathcal{N}_{i}^{t,s}=\{\mathbf{d}_j^t\in\mathcal{D}_t:\|\mathbf{d}_j^t-\mathbf{x}_i^{(s)}\|<d_{\max}\}$. We define the unnormalized Gaussian weights and the resulting observation force as
\begin{align}
\omega_{ij}^{t,s}
&=\exp\left(-\frac{\|\mathbf{d}_j^t-\mathbf{x}_i^{(s)}\|^2}{2\sigma^2}\right),
\label{eq:obs_weight_method}\\
\mathbf{F}_{i}^{\rm obs,(s)}
&={\rm vis}_i^{(s)}k_{\rm obs}
\sum_{\mathbf{d}_j^t\in\mathcal{N}_{i}^{t,s}}
\omega_{ij}^{t,s}\left(\mathbf{d}_j^t-\mathbf{x}_i^{(s)}\right).
\label{eq:obs_force_method}
\end{align}
Here $k_{\rm obs}>0$ controls the strength of the observation force, implemented as $k_{\rm obs}=\exp(\kappa_{\rm obs})$. $\sigma$ is the Gaussian kernel scale, and $d_{\max}$ truncates far-away observations. 
The observation force is added to the spring--mass simulator together with the internal spring--dashpot forces and action-induced external forces, producing the corrected state $\hat{\mathbf{X}}_t$ through physical integration.
Visibility and observation forces are recomputed at each simulation substep within a frame.


\subsection{physics-aware 4D Mapping}
\label{sec:physics_4d_mapping}

After tracking, we perform 4D mapping to update the static background field, the spring--mass physical model, and the dynamic object Gaussian field.
For the static branch, we follow the Gaussian map expansion and local-window optimization of WildGS-SLAM~\cite{zheng2025wildgs}, but replace its learned uncertainty map with our semantic background mask $M_t^{\rm bg}$.
This uses the same static/dynamic partition as tracking and keeps moving object pixels in the dynamic branch. New background Gaussians are inserted from unmapped static pixels and optimized with masked RGB-D rendering losses over a local keyframe window.

For the dynamic branch, we update the physical parameters $\boldsymbol{\theta}=\{m_i,k_{ij},\gamma_{ij},\ell_{ij}\}$ (mass, stiffness, damping, and rest length) based on the geometry anchors from observation-corrected states, while we transform the dynamic Gaussians with the corrected physical surface and refine them, together with our proposed surface-alignment regularization as follows.

\noindent\textbf{Physical parameter update.}
The corrected states from tracking provide stable anchors for updating physical parameters.
Over the full observation window $\mathcal{W}_t$ used for reconstruction, 
the physical parameters are updated by minimizing
\begin{equation}
    \mathcal{L}_{\rm geo}
    =
    \sum_{\tau\in\mathcal{W}_t}
    \mathcal{L}_{\rm CD}
    \left(
    \operatorname{Pos}
    \left(
    \hat{\mathbf{X}}_{\tau}(\boldsymbol{\theta})
    \right),
    \mathcal{D}_{\tau}
    \right)
    +
    \lambda_{\theta}
    \left\|
    \boldsymbol{\theta}
    -
    \boldsymbol{\theta}_{t-1}
    \right\|_2^2 .
    \label{eq:physical_update}
\end{equation}
Here $\tau$ indexes frames in $\mathcal{W}_t$, $\mathcal{D}_{\tau}$ is the observed object point cloud, and $\boldsymbol{\theta}_{t-1}$ is the previous parameter estimate. $\operatorname{Pos}(\cdot)$ extracts surface node positions. The Chamfer distance $\mathcal{L}_{\rm CD}$ compares the simulated and observed point sets without requiring point-to-point correspondences; $\lambda_{\theta}$ weights the penalty on abrupt parameter changes.
Since each rollout starts from an observation-corrected state, this update reduces the coupling between tracking drift and physical model error.

\noindent\textbf{physics-aware Gaussian transformation.}
Given the corrected physical state, we update the surface mesh $\mathcal{M}_t$ and transform dynamic Gaussians using the binding set 
$\mathcal{B}=\{(q(n),\boldsymbol{\beta}_n,h_n)\}_{g_n\in\mathcal{G}^{\rm obj}_{\rm c}}$ 
defined in Eq.~\eqref{eq:gaussian_binding}.
Specifically, the new Gaussian centers are computed as
\begin{equation}
    \bar{\boldsymbol{\mu}}_n^t
    =
    \sum_{r=1}^{3}
    \beta_{nr}
    \mathbf{p}_{q(n),r}^{t}
    +
    h_n
    \mathbf{n}_{q(n)}^{t}.
    \label{eq:physics_transport}
\end{equation}
The physics-aware transformation keeps Gaussians attached to the deforming physical surface.
Different from Linear Blend Skinning (LBS)~\cite{magnenat1988joint,lewis2000pose,loper2015smpl,jiang2025phystwin} or control-point-based interpolation~\cite{li20254dgsslam}, our method directly transforms Gaussians with the deforming physical surface without repeated K-Nearest Neighbors (KNN) search or local interpolation during 4D mapping.

\noindent\textbf{Surface-alignment regularization of dynamic Gaussians.}
After transformation, we refine the dynamic Gaussian attributes with the object rendering loss $\mathcal{L}_{\rm render}^{\rm obj}$, a weighted combination of $L_1$ and D-SSIM color terms over object pixels following PhysTwin~\cite{jiang2025phystwin}.
Our key proposed constraint is the surface-alignment regularizer:
\begin{equation}
    \mathcal{L}_{\rm surf}
    =
    \frac{1}{|\mathcal{G}^{\rm obj}_t|}
    \sum_{g_n\in\mathcal{G}^{\rm obj}_t} \frac{1}{\bar{s}_n+\delta}
    \rho
    \left(
    (\mathbf{n}_{q(n)}^{t})^{\top}\boldsymbol{\mu}_n^t
    +
    d_{q(n)}^{t}
    \right),
    \label{eq:surface_alignment}
\end{equation}
Here $\mathbf{n}_{q(n)}^{t}$ is the unit face normal and $d_{q(n)}^{t}$ is the plane offset such that 
$(\mathbf{n}_{q(n)}^{t})^{\top}\mathbf{x}+d_{q(n)}^{t}=0$, and $\rho(r)={(r^2+\epsilon_\rho^2)}^{1/2}$ is a Charbonnier penalty with a small constant $\epsilon_\rho>0$ for smoothness near zero and numerical stability.
The argument of $\rho$ is the signed distance from the Gaussian center to the plane of its associated mesh face. $\bar{s}_n$ is the mean Gaussian scale and $\delta>0$ prevents division by zero, giving larger Gaussians lower weights.
This soft penalty suppresses normal drift while allowing tangential adjustments for rendering quality. The stored offset $h_n$ transports Gaussians with mesh deformation and is not forced to zero.

The surface-alignment term $\lambda_{\rm surf}\mathcal{L}_{\rm surf}$ in $\mathcal{L}_{\rm map}^{\rm obj}$ penalizes normal drift during Gaussian appearance optimization, while the surface binding allows occluded regions to inherit physically plausible motion from the spring--mass model. The overall object mapping loss is
\begin{equation}
    \mathcal{L}_{\rm map}^{\rm obj}
    =
    \lambda_{\rm geo}
    \mathcal{L}_{\rm geo}
    +
    \lambda_{\rm render}
    \mathcal{L}_{\rm render}^{\rm obj}
    +
    \lambda_{\rm surf}
    \mathcal{L}_{\rm surf}.
    \label{eq:dynamic_mapping_objective}
\end{equation}
This loss jointly updates the physical parameters and the dynamic Gaussian field while keeping the Gaussian centers aligned with the evolving physical surface.

\subsection{Future action-conditioned Prediction}
\label{sec:future_prediction}

After reconstruction, the dynamic object field can be rolled forward beyond observed frames, similar to \cite{jiang2025phystwin}. Future control-point trajectories are given action inputs; future object RGB-D observations are used only for evaluation, without further state correction or parameter optimization.
Given the latest corrected state $\hat{\mathbf{X}}_t$, updated physical parameters $\boldsymbol{\theta}_t$, and future actions $\{\mathbf{a}_{t+\ell}\}_{\ell=0}^{H-1}$, we disable the observation force because we should not know the future obervation during future prediction, set $\mathbf{X}_{t}^{\rm pred}=\hat{\mathbf{X}}_t$, and autoregressively roll out the spring--mass simulator defined in Sec.~\ref{sec:Preliminary} as $\mathbf{X}_{t+\ell+1}^{\rm pred}=f_{\boldsymbol{\theta}_t,\mathcal{H}}(\mathbf{X}_{t+\ell}^{\rm pred},\mathbf{a}_{t+\ell})$ for $\ell=0,\dots,H-1$. For each predicted state $\mathbf{X}_{t+\ell}^{\rm pred}$, we apply the physics-aware Gaussian transformation in Eq.~\eqref{eq:physics_transport} with the predicted surface vertices and normals to obtain the predicted Gaussian centers $\bar{\boldsymbol{\mu}}_n^{t+\ell,{\rm pred}}$. The predicted object field $\bar{\mathcal{G}}_{t+\ell}^{\rm obj,pred}$ keeps the optimized Gaussian attributes from $\mathcal{G}^{\rm obj}_t$ and replaces their centers with the predicted centers.
Rendering $\mathcal{G}^{\rm bg}\cup\bar{\mathcal{G}}_{t+\ell}^{\rm obj,pred}$ from a given query camera pose yields future appearance and geometry predictions. Thus, the reconstructed field represents observed frames and supports physically grounded future prediction.

\section{Experiments}
\label{sec:experiments}

\begin{figure*}[t]
    \centering
    \includegraphics[width=1.00\textwidth]{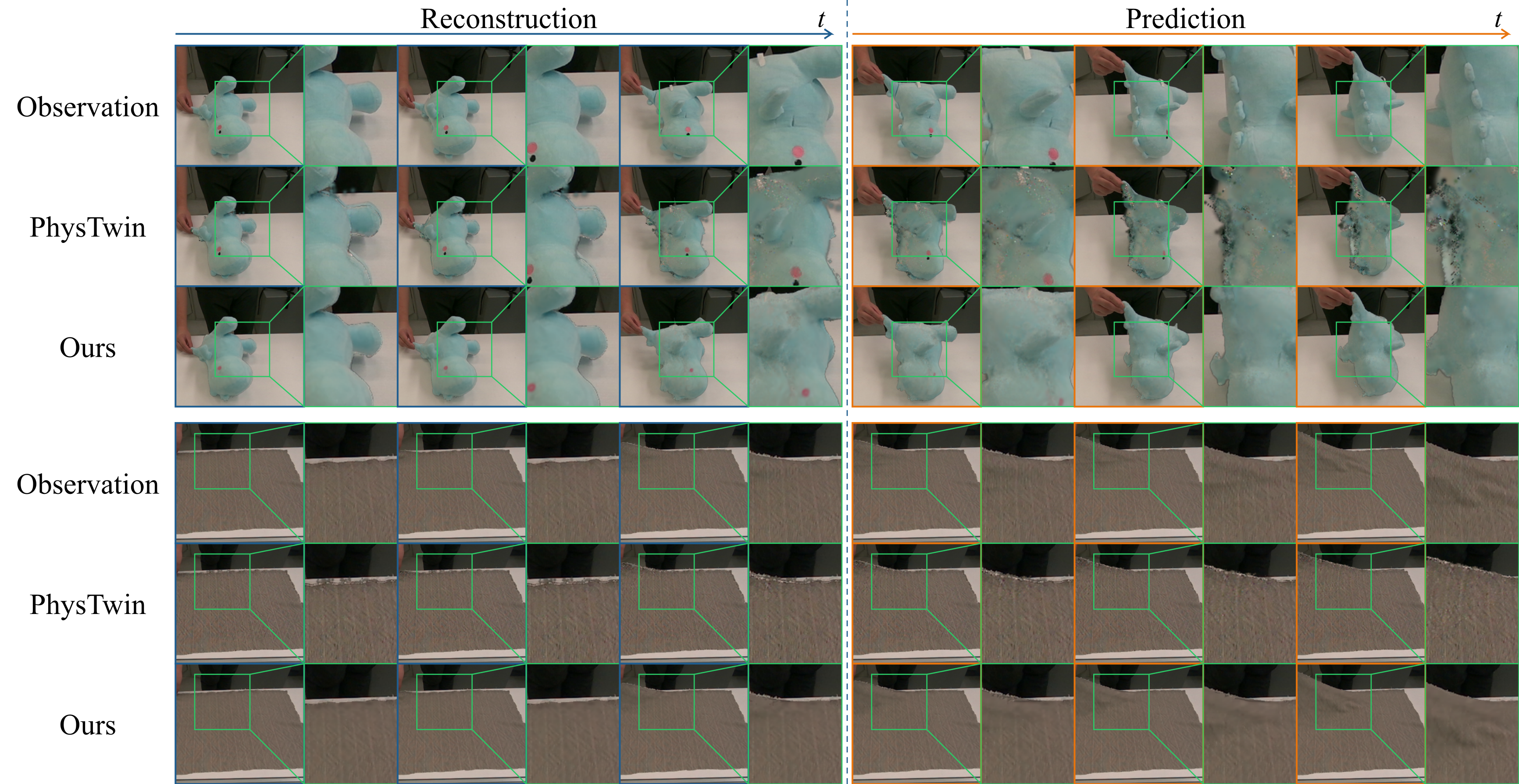}
    \vspace{-6mm}
      \caption{
\textbf{Qualitative Results Compared with PhysTwin.}
Ours produces renderings that better align with the observations, with sharper appearance and more accurate silhouettes. Green boxes and connecting lines indicate magnified regions; blue and orange borders distinguish reconstruction and prediction, respectively.}
    \label{fig:qual_phystwin}
\end{figure*}


\begin{table*}[t]
\caption{Comparison with state-of-the-art methods on the PhysTwin benchmark.}
\label{tab:comparison_phystwin}
\centering
\setlength{\tabcolsep}{4pt}
\resizebox{\textwidth}{!}{
\begin{tabular}{lcccccc|cccccc}
\toprule
\multirow{2}{*}{Method}
& \multicolumn{6}{c|}{Reconstruction}
& \multicolumn{6}{c}{Prediction} \\
\cmidrule(lr){2-7}
\cmidrule(lr){8-13}
& CD$\downarrow$ & Track$\downarrow$ & IoU(\%)$\uparrow$ & PSNR$\uparrow$ & SSIM$\uparrow$ & LPIPS$\downarrow$
& CD$\downarrow$ & Track$\downarrow$ & IoU(\%)$\uparrow$ & PSNR$\uparrow$ & SSIM$\uparrow$ & LPIPS$\downarrow$ \\
\midrule

Spring-Gau~\cite{zhong2024springgaus}
& 0.041 & 0.050 & 57.6 & 23.445 & 0.928 & 0.102
& 0.062 & 0.094 & 46.4 & 22.488 & 0.924 & 0.113 \\

GS-Dynamics~\cite{zhang2025gsdynamics}
& 0.014 & 0.022 & 72.1 & 26.260 & 0.940 & 0.052
& 0.041 & 0.070 & 49.8 & 22.540 & 0.924 & 0.097 \\

PhysTwin~\cite{jiang2025phystwin}& \textbf{0.005} & \textbf{0.009} & 84.4 & 28.214 & 0.945 & 0.034& 0.012 & \textbf{0.022} & 72.5 & 25.617 & 0.941 & 0.055 \\

Ours
& \textbf{0.005} & \textbf{0.009} & \textbf{91.9} & \textbf{31.140} & \textbf{0.980} & \textbf{0.024}
& \textbf{0.011} & \textbf{0.022} & \textbf{75.9} & \textbf{28.430} & \textbf{0.961} & \textbf{0.049} \\

\bottomrule
\end{tabular}
}
\end{table*}

\begin{figure*}[t]
    \centering
\includegraphics[width=1.0\textwidth]{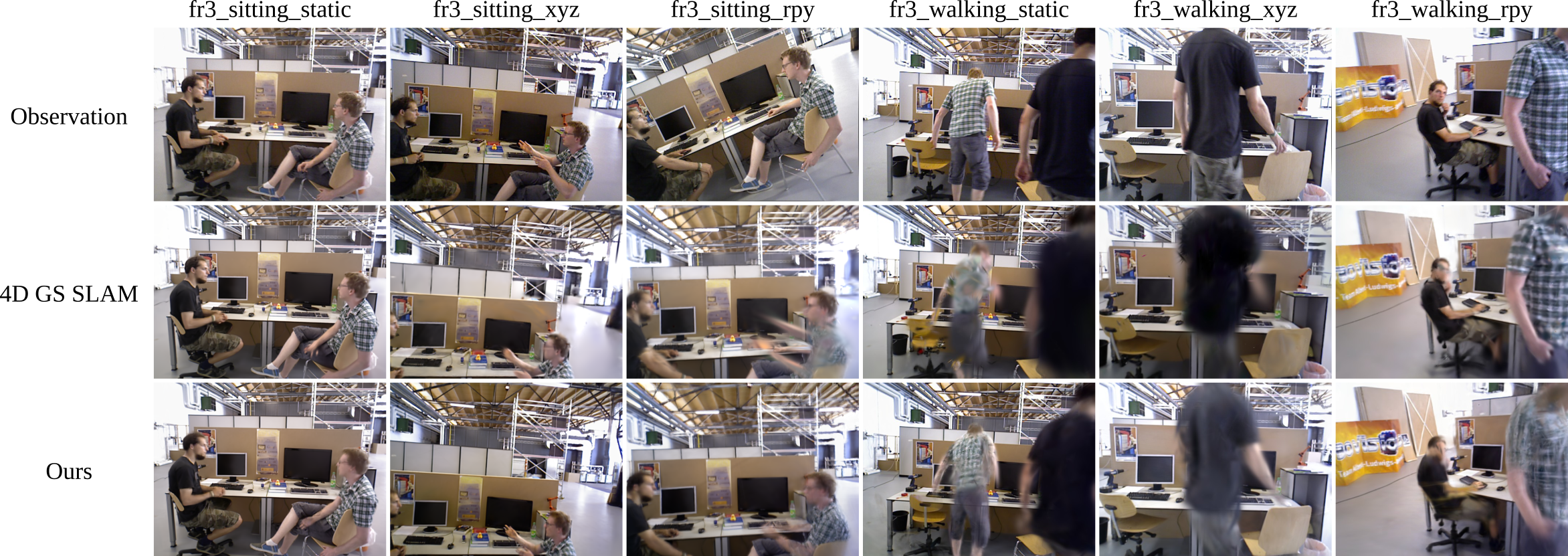}
    \vspace{-6mm}
\caption{
\textbf{Visual comparison of rendering results on the TUM RGB-D dataset.}
}
\label{fig:tum_rendering_vis}
    \label{fig:qual_4DGS}
\end{figure*}

\begin{table*}[t]
\caption{Trajectory ATE RMSE [cm] $\downarrow$ on the TUM RGB-D sequences.}
\label{tab:tum_ate}
\centering
\setlength{\tabcolsep}{6pt}
\resizebox{0.9\linewidth}{!}{
\begin{tabular}{lccccccc}
\toprule
Method & fr3/sit\_st & fr3/sit\_xyz & fr3/sit\_rpy & fr3/walk\_st & fr3/walk\_xyz & fr3/walk\_rpy & Avg. \\
\midrule
RoDyn-SLAM~\cite{jiang2024rodynslam}
& 1.5 & 5.6 & 5.7 & 1.7 & 8.3 & 8.1 & 5.1 \\
MonoGS~\cite{matsuki2024gaussian}
& \textbf{0.48} & 1.7 & 6.1 & 21.9 & 30.7 & 34.2 & 15.8 \\
SplaTAM~\cite{keetha2024splatam}
& 0.52 & 1.5 & 11.8 & 83.2 & 134.2 & 142.3 & 62.2 \\
4DGS-SLAM~\cite{li20254dgsslam}
& 0.58 & 2.9 & 2.6 & 0.52 & 2.1 & \textbf{2.6} & 1.8 \\
Ours
& 0.53 & \textbf{0.87} & \textbf{2.29} & \textbf{0.50} & \textbf{1.23} & 3.13 & \textbf{1.43} \\
\bottomrule
\end{tabular}
}
\end{table*}

\begin{table*}[t!]
\caption{Rendering quality on the TUM RGB-D FR3 sequences.}
\label{tab:tum_rendering}
\centering
\resizebox{0.9\textwidth}{!}{
\begin{tabular}{llccccccc}
\toprule
\textbf{Method} & \textbf{Metric}
& \textbf{fr3/sit\_st}
& \textbf{fr3/sit\_xyz}
& \textbf{fr3/sit\_rpy}
& \textbf{fr3/walk\_st}
& \textbf{fr3/walk\_xyz}
& \textbf{fr3/walk\_rpy}
& \textbf{Avg.} \\
\midrule

\multirow{3}{*}{MonoGS~\cite{matsuki2024gaussian}}
& PSNR$\uparrow$ 
& 19.95 & 23.92 & 16.99 & 16.47 & 14.02 & 15.12 & 17.74 \\
& SSIM$\uparrow$
& 0.739 & 0.803 & 0.572 & 0.604 & 0.436 & 0.497 & 0.608 \\
& LPIPS$\downarrow$
& 0.213 & 0.182 & 0.405 & 0.355 & 0.581 & 0.560 & 0.382 \\

\midrule

\multirow{3}{*}{Gaussian-SLAM~\cite{yugay2023gaussianslam}}
& PSNR$\uparrow$
& 18.57 & 19.22 & 16.75 & 14.91 & 14.67 & 14.50 & 16.43 \\
& SSIM$\uparrow$
& 0.848 & 0.796 & 0.652 & 0.607 & 0.483 & 0.467 & 0.642 \\
& LPIPS$\downarrow$
& 0.291 & 0.326 & 0.521 & 0.489 & 0.626 & 0.630 & 0.480 \\

\midrule

\multirow{3}{*}{SplaTAM~\cite{keetha2024splatam}}
& PSNR$\uparrow$
& 24.12 & 22.07 & 19.97 & 16.70 & 17.03 & 16.54 & 19.40 \\
& SSIM$\uparrow$
& \textbf{0.915} & \textbf{0.879} & \underline{0.799} 
& 0.688 & 0.650 & 0.635 & 0.757 \\
& LPIPS$\downarrow$
& \textbf{0.101} & \underline{0.163} & \textbf{0.205}
& 0.287 & 0.339 & 0.353 & 0.241 \\

\midrule

\multirow{3}{*}{4DGS-SLAM~\cite{li20254dgsslam}}
& PSNR$\uparrow$
& \textbf{27.68} & \underline{24.37} & \underline{20.71}
& \textbf{22.99} & 19.83 & \underline{19.22}
& \textbf{22.46} \\
& SSIM$\uparrow$
& 0.892 & 0.822 & 0.746
& \textbf{0.820} & \underline{0.730} & \underline{0.708}
& \underline{0.786} \\
& LPIPS$\downarrow$
& \underline{0.116} & 0.179 & 0.265
& \textbf{0.195} & \underline{0.281} & \underline{0.337}
& \underline{0.228} \\

\midrule

\multirow{3}{*}{\textbf{Ours}}
& PSNR$\uparrow$
& 24.68 & \textbf{24.57} & \textbf{22.96}
& \underline{21.45} & \textbf{20.17} & \textbf{20.02}
& \underline{22.31} \\
& SSIM$\uparrow$
& 0.878 & \underline{0.866} & \textbf{0.816}
& \underline{0.810} & \textbf{0.769} & \textbf{0.743}
& \textbf{0.814} \\
& LPIPS$\downarrow$
& 0.127 & \textbf{0.145} & \underline{0.210}
& \underline{0.209} & \textbf{0.254} & \textbf{0.309}
& \textbf{0.209} \\

\bottomrule
\end{tabular}
}
\end{table*}

We evaluate Physical SLAM from three complementary perspectives.
First, we follow the PhysTwin benchmark protocol~\cite{jiang2025phystwin} to evaluate deformable-object reconstruction and future prediction.
Second, we follow the dynamic Gaussian SLAM setting of 4DGS-SLAM~\cite{li20254dgsslam} on the TUM RGB-D dataset~\cite{sturm2012benchmark} to evaluate camera tracking and dynamic scene reconstruction.
Third, we collect a moving-camera manipulation dataset to evaluate reconstruction quality when both the camera and the deformable object are moving.

\subsection{Reconstruction and Prediction on the PhysTwin Dataset}
\label{sec:exp_phystwin}

\noindent\textbf{Setting.}
We use the same dataset, data processing methods, train/test split, tasks, and evaluation metrics as PhysTwin~\cite{jiang2025phystwin}.
The benchmark contains RGB-D videos of deformable objects under human interaction, including stuffed toys, ropes, cloth-like objects, and packages.
Following PhysTwin, we evaluate two tasks:
\emph{Reconstruction}, where the system reconstructs observed interactions, and
\emph{Prediction}, where the system predicts unseen future frames under given future control-point actions. We report the same 3D metrics, including single-direction Chamfer Distance and tracking error, and 2D metrics, including IoU, PSNR, SSIM, and LPIPS.

\noindent\textbf{Results.}
The Reconstruction column in Table~\ref{tab:comparison_phystwin} shows that our method achieves strong reconstruction performance. Compared with PhysTwin, our method matches the best 3D geometry metrics, while substantially improving 2D alignment and rendering quality. This indicates that the proposed physics-aware Gaussian transformation and surface-alignment regularization preserve physical consistency while producing sharper and better-aligned renderings.
The Prediction column in Table~\ref{tab:comparison_phystwin} further shows that our method improves over PhysTwin in CD, IoU, PSNR, SSIM, and LPIPS, while matching the best Track Error. These results suggest that the progressively updated spring--mass model captures physical dynamics that generalize to future states, while the surface-bound Gaussian field transfers the predicted deformation into coherent appearance and geometry. Fig.~\ref{fig:qual_phystwin} also shows that our method reconstructs and predicts more plausible object deformation by binding dynamic Gaussians to the deforming physical surface.

\begin{table*}[t]
\caption{Quantitative results on our moving-camera manipulation dataset.}
\label{tab:ours_dataset_mapping}
\centering
\setlength{\tabcolsep}{8pt}
\resizebox{0.7\textwidth}{!}{
\begin{tabular}{lccc|ccc}
\toprule
\multirow{2}{*}{Method}
& \multicolumn{3}{c|}{Reconstruction}
& \multicolumn{3}{c}{Prediction} \\
\cmidrule(lr){2-4}
\cmidrule(lr){5-7}
& PSNR$\uparrow$ & SSIM$\uparrow$ & LPIPS$\downarrow$
& PSNR$\uparrow$ & SSIM$\uparrow$ & LPIPS$\downarrow$ \\
\midrule
4DGS-SLAM~\cite{li20254dgsslam}
& 22.65 & 0.774 & \textbf{0.251}
& -- & -- & -- \\
Ours
& \textbf{24.20} & \textbf{0.817} & 0.295
& \textbf{24.09} & \textbf{0.831} & \textbf{0.263} \\
\bottomrule
\end{tabular}
}
\end{table*}


\begin{table*}[t]
\caption{Ablation study on the PhysTwin benchmark.}
\label{tab:ablation_phystwin}
\centering
\setlength{\tabcolsep}{3pt}
\resizebox{\textwidth}{!}{
\begin{tabular}{lcccccc|cccccc}
\toprule
\multirow{2}{*}{Variant}
& \multicolumn{6}{c|}{Reconstruction}
& \multicolumn{6}{c}{Prediction} \\
\cmidrule(lr){2-7}
\cmidrule(lr){8-13}
& CD$\downarrow$ & Track$\downarrow$ & IoU(\%)$\uparrow$ & PSNR$\uparrow$ & SSIM$\uparrow$ & LPIPS$\downarrow$
& CD$\downarrow$ & Track$\downarrow$ & IoU(\%)$\uparrow$ & PSNR$\uparrow$ & SSIM$\uparrow$ & LPIPS$\downarrow$ \\
\midrule

Full
& \textbf{0.005} & \textbf{0.009} & \textbf{91.9} & \textbf{31.140} & \textbf{0.980} & \textbf{0.024}
& \textbf{0.011} & \textbf{0.022} & \textbf{75.9} & \textbf{28.430} & \textbf{0.961} & \textbf{0.049} \\

w/o obs-force
& 0.005 & 0.010 & 83.5 & 29.650 & 0.958 & 0.044
& 0.012 & 0.022 & 68.2 & 25.120 & 0.945 & 0.064 \\

w/o surface align loss
& 0.005 & 0.009 & 86.0 & 30.353 & 0.964 & 0.026
& 0.011 & 0.022 & 69.7 & 25.965 & 0.947 & 0.051 \\


\bottomrule
\end{tabular}
}
\end{table*}
\subsection{Dynamic Gaussian SLAM Evaluation on TUM RGB-D}
\label{sec:exp_tum}

\noindent\textbf{Setting.}
We evaluate camera tracking and reconstruction quality on TUM RGB-D sequences following the 4DGS-SLAM protocol~\cite{li20254dgsslam}.
We report Absolute Trajectory Error (ATE) RMSE for camera tracking, and PSNR, SSIM, and LPIPS for reconstruction quality.

\noindent\textbf{Results.}
As shown in Table~\ref{tab:tum_ate}, our method achieves the best average trajectory accuracy among all compared methods. Compared with static Gaussian SLAM methods such as MonoGS and SplaTAM, our method is substantially more robust on the walking sequences, where large foreground motion causes severe tracking degradation for methods that assume a mostly static scene. Compared with 4DGS-SLAM, our method achieves the best overall average. This indicates that separating the static background for camera tracking can improve camera localization robustness.
Table~\ref{tab:tum_rendering} shows
our method improves average SSIM and LPIPS over 4DGS-SLAM, with slightly lower PSNR (22.31 versus 22.46 dB).
Our method achieves the best PSNR, SSIM, and LPIPS on fr3/walk\_xyz and fr3/walk\_rpy. This suggests that the physics-aware dynamic branch helps a lot in reducing inconsistent dynamic reconstruction.
As visualized in Fig.~\ref{fig:tum_rendering_vis}, compared with 4DGS-SLAM, which models dynamic regions mainly through optical-flow-based constraint, our physics-aware Gaussian transformation produces more stable and sharper dynamic regions.
Benefiting from mesh binding and physical surface motion, our method better preserves dynamic object structure and reduces motion-induced artifacts, especially in highly dynamic walking sequences.

\subsection{Evaluation on Moving-camera Manipulation Dataset}
\label{sec:exp_ours_dataset}

\noindent\textbf{Dataset.}
To further evaluate our system, we collect a new RGB-D dataset where a moving camera observes a human manipulating deformable objects, which enables our method to make future predictions with external actions like PhysTwin~\cite{jiang2025phystwin}. The sequence contains 207 frames: the first 124 are used for reconstruction and the remaining 83 for action-conditioned prediction. We compare primarily with the closest method, 4DGS-SLAM~\cite{li20254dgsslam}, which is the 4D Gaussian SLAM baseline but does not model physical dynamics.

\noindent\textbf{Results.} 
As shown in Table~\ref{tab:ours_dataset_mapping}, since 4DGS-SLAM is designed for reconstruction and does not include a physical prediction module, the prediction metrics are also unavailable. For reconstruction, our method improves PSNR (24.20 versus 22.65 dB) and SSIM (0.817 versus 0.774) over 4DGS-SLAM, while LPIPS is a little worse (0.295 versus 0.251). These results show improved pixelwise and structural similarity, despite a higher perceptual error. Moreover, our method can further roll out future states, demonstrating its ability to not only reconstruct the observed scene but also capture the underlying dynamics for action-conditioned future prediction.

\subsection{Ablation Study}
\label{sec:exp_ablation}

We conduct individual ablations on the PhysTwin benchmark to analyze the contribution of each component.
\textbf{w/o obs-force} removes observation-force correction.
\textbf{w/o surface align loss} removes the surface-alignment regularizer during dynamic Gaussian optimization.

Table~\ref{tab:ablation_phystwin} shows the effectiveness of each component. Removing the observation force weakens both reconstruction and prediction quality, especially IoU and PSNR. These results support our observation-force design for coupling visual observations with physical dynamics: geometry-based forces correct the surface state that drives dynamic Gaussians. The prediction gains persist after observation forces are disabled, suggesting that reconstruction-stage correction also benefits subsequent action-conditioned rollout.
Removing the surface-alignment loss reduces IoU and rendering quality in both tasks. The rendering gains support the contribution of surface alignment to maintaining appearance--geometry consistency during Gaussian refinement. This is consistent with the regularizer suppressing normal drift while retaining tangential adjustment, helping the refined Gaussians stay aligned with the physical surface that drives their motion.

\raggedbottom
\section{Conclusion}
\label{sec:conclusion}

We presented Physical SLAM, a physics-aware 4D Gaussian reconstruction and prediction method that couples dynamic Gaussians with a physical model.
Unlike existing dynamic Gaussian SLAM systems that primarily model visual motion, our method reconstructs the physical state and parameters of deformable objects from streaming RGB-D observations. 
Limitation and Future Work: Our current system still assumes a fixed spring--mass topology. Future work will relax these assumptions by supporting topology-changing deformation to reconstruct tearable objects.

\bibliographystyle{IEEEtran}
\bibliography{main}






\end{document}